\documentclass[11pt]{article}

\usepackage[]{ACL2023}

\usepackage{times}
\usepackage{latexsym}
\usepackage{amsmath}
\usepackage[ruled,linesnumbered]{algorithm2e}
\usepackage{graphicx}
\usepackage{enumitem}
\usepackage{booktabs}
\usepackage{multirow}
\usepackage{makecell}

\usepackage{amsthm}

\newtheorem{observation}{Observation}
\usepackage[inkscapelatex=false]{svg}
\usepackage{stfloats}
\usepackage{float}
\usepackage[T1]{fontenc}

\usepackage[utf8]{inputenc}

\usepackage{microtype}

\usepackage{inconsolata}

\title{Injecting Knowledge into Biomedical Pre-trained Models\\ via Polymorphism and Synonymous Substitution}

\author{Hongbo Zhang \and Xiang Wan \and Benyou Wang\thanks{Benyou Wang is the corresponding author} \\
        School of Data Science, The Chinese University of Hong Kong, Shenzhen \\ Shenzhen Research Institute of Big Data \\
        \texttt{hongboz183@gmail.com}, \texttt{wangbenyou@cuhk.edu.cn} \\}

\begin{document}
\maketitle
\begin{abstract}

Pre-trained language models (PLMs) were considered to be able to store relational knowledge present in the training data. However, some relational knowledge seems to be discarded unsafely in PLMs due to \textbf{report bias}: low-frequency relational knowledge might be underexpressed compared to high-frequency one in PLMs. This gives us a hint that relational knowledge might not be redundant to the stored knowledge of PLMs, but rather be complementary.
To additionally inject relational knowledge into PLMs, we propose a simple-yet-effective approach to inject relational knowledge into PLMs, which is inspired by three observations (namely, polymorphism, synonymous substitution, and association). In particular, we switch entities in the training corpus to related entities (either hypernyms/hyponyms/synonyms, or arbitrarily-related concepts).
Experimental results show that the proposed approach could not only better capture relational knowledge, but also improve the performance in various biomedical downstream tasks. Our model is available in \url{https://github.com/StevenZHB/BioPLM_InjectingKnowledge}.

\end{abstract}

\section{Introduction}
Transformers pre-trained on a large amount of unlabelled corpus~\cite{devlin2018bert,qiu2020pre} have been claimed to store many relational knowledge~\cite{petroni2019language,Bouraoui_Camacho-Collados_Schockaert_2020}. 
However, our pilot study and many existing works find that PLMs are insensitive to capture low-frequency relational knowledge (\textit{a.k.a.}, report bias \cite{gordon2013reporting,shwartz2020neural}). It is not guaranteed that PLMs could properly remember even high-frequency knowledge \cite{petroni2019language,cao2021knowledgeable}. Therefore, relational knowledge might not be redundant to the stored knowledge of PLMs, but rather be complementary, see our pilot study in \S\ref{sec:complementarity}.

This work aims to inject relational knowledge into PLMs.  We select biomedical relational knowledge as a case study. It is more challenging in biomedicine since there usually exists  1) multiple synonyms due to the non-standardized terminology and 2) the hierarchy of biomedical concepts. As a starter, we will first introduce some observations.

\begin{figure}[htb]
\centering \small
\vspace{-35pt}
\includegraphics[width=\columnwidth]{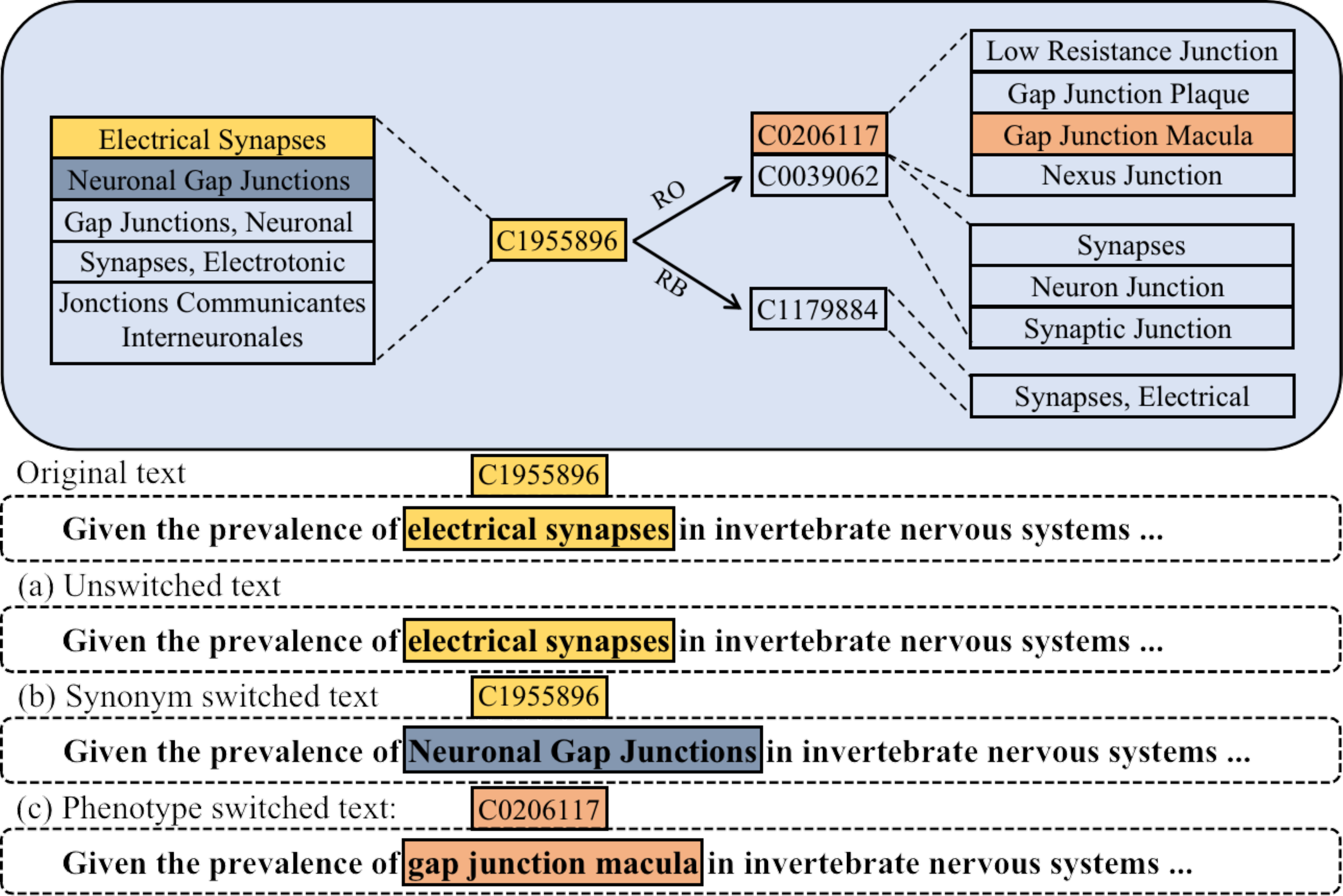}
\caption{A switching example from the corpus. The codes beginning with 'C' are concept IDs in UMLS. Three switching methods are shown.
}
\label{intro_fig}
\end{figure}

\begin{observation}
\label{ob:Polymorphism}
\textbf{Polymorphism},  in biology, is the occurrence of two or more clearly different morphs or forms, in the population of a species. In UMLS, there is a hierarchy between concepts, due to the prevalence of hypernyms and hyponyms. The hyponym of each concept can usually inherit the features of its parent concepts.


\end{observation}

\begin{observation}
\label{ob:Synonymous substitution}
\textbf{Synonymous substitution}  is the evolutionary substitution of one base for another in an exon of a gene coding for a protein that does not modify the produced amino acid sequence. Here, we found that concepts usually have a collection of synonyms that corresponds to the same ID. 
\end{observation}

Note that in Observation~\ref{ob:Polymorphism} and ~\ref{ob:Synonymous substitution}, replacement is safe since it does not change the semantic aspect of the text.
To be more general, we introduce Observation~\ref{ob:Association} which might introduce some unexpected semantic vibration but offers more generality. 

\begin{observation}
\label{ob:Association}
\textbf{Association} is to replace a target entity with its associated entity.
Typically this might lightly modify texts semantically but the association between concepts is augmented.
\end{observation}

However, current PLMs are insensitive to polymorphism and synonymous substitution, see \S\ref{sec:deficiency}. To compensate for the above deficiency, we propose a simple-yet-effective approach to inject relational knowledge into PLMs without modifying the model structure: switching entity pairs with different relations including hypernym, hyponym, synonym, etc, as shown in Fig.~\ref{intro_fig}.
In detail, we first sample some target concepts in the training corpus and then randomly replace them with their relevant concepts that have specific relations (e.g., hypernym, hyponym, synonym) with the target concepts, probabilities of which are dependent on the relational category. 
Our experimental results illustrate that our proposed approach could not only better capture relational knowledge, but also improve various biomedical downstream tasks.

 




\section{A Pilot Study}
\subsection{Complementarity between KB  and PLMs }
\label{sec:complementarity}

\begin{table}[t]
\vspace{-5pt}
    \centering \footnotesize
    \begin{tabular}{cccc}
        \toprule
        Subject Frequency & Low & Medium & High \\
        \midrule
        Acc@1 & 7.3 & 7.6 & 9.6 \\
        \midrule
        Data & UMLS & UMLS-Syn & UMLS-Hyp \\
        \midrule
        Acc@1 & 5.15 & 4.61 & 4.3 \\
        \bottomrule
    \end{tabular}
    \caption{Knowledge probing  on PubMedBert. The top is for with different subject frequency and the bottom is  replacing  subjects  with synonyms or hyponyms. }
    \label{tab: quantitative analysis}
\end{table}

\label{sec:pilot}
Quantitative analysis is shown in Tab.~\ref{tab: quantitative analysis}, we sample three groups of BioLAMA dataset according to the occurrence frequency of subjects in our corpus and probe knowledge stored in Biomedical Bert. It shows that \textbf{PLMs are vulnerable to report bias}: PLMs capture better knowledge that is related to high-frequency entities than that of low-frequency entities. Unlike PLMs that are biased to the entity frequency, triplets in knowledge bases are not vulnerable since knowledge triplets are equal no matter whether the corresponding entities are high-frequency or low-frequency. Therefore, knowledge triplets might be complementary to PLMs, especially for knowledge with low-frequency entities.   See App.~\ref{sec:example_complementary} for a concrete example.

\subsection{Deficiency of PLMs in Polymorphism and Synonymous substitution}
\label{sec:deficiency}
As shown in Tab.~\ref{tab: quantitative analysis}, a simple experiment was carried out to probe the knowledge of which subjects are replaced by synonyms or hyponyms. 
Note that even if the subjects are replaced, the knowledge meaning would generally not be changed and it should predict the identical objects for the masked position. Experimental result shows that the performance in BioLAMA is largely decreased, demonstrates that PLMs are vulnerable to polymorphism and synonymous substitution.


\section{Methodology}

\subsection{Formal Definition}

For a given knowledge base with triplets $ \mathcal{T} = \{s, r , o \}$, each of which has a \textit{subject} entity $s$,  an \textit{object} entity $o$, and the \textit{relation}  $r$ among them. We split these triplets into many groups according to the relations; subject-object entity pairs in each group are defined as a set  $\Theta$. It results in the total set $\Theta = [\Theta_{r_1},  \Theta_{r_2}, \cdots,  \Theta_{r_K}  ]$, where $\Theta_{r}$ is the  subject-object entity set for a specific relation $r$, namely, $\Theta_{r}  = \{ s_i,  o_i \} $ and $(s_i, r, o_i) \in \mathcal{T}$.

In UMLS, there are synonymous entities within a same concept ID, we denote these entities to have a Synonymous Relation (\textbf{SR}); this is generally accurate thanks to human annotation. Furthermore, there are 13 other relations in UMLS~\footnote{\footnotesize \url{https://www.nlm.nih.gov/research/UMLS/knowledge_sources/metathesaurus/release/abbreviations.html}, see App.~\ref{app:Relation Details} for more details about relations in UMLS.} including \textbf{CHD} that indicates a child relationship in a Metathesaurus source vocabulary.
The observation~\ref{ob:Polymorphism} (polymorphism) and  \ref{ob:Synonymous substitution} (synonymous substitution) suggest that \textit{replacement from the subject entity to object entity in $\Theta_\textrm{CHD}$ and $\Theta_\textrm{SR}$ is generally valid};  it augments PLMs with better perception of polymorphism and synonymous substitution.

Based on Observation~\ref{ob:Association} (association), one could associate an entity with another relevant entity if they are with either a strong relation (e.g. \textbf{CHD} and \textbf{SR}) or a weak one. This could implicitly augment PLMs with any relations defined in knowledge bases. We denote the relation set as $\mathcal{R}$ including \textbf{SR} and other 13 relations in UMLS.

\subsection{Entity Switching}
\label{sec:entity switching}
We employ entity switching in pre-training corpus to implicitly inject concepts and relations into PLMs. The switching process can be illustrated in Algo.~\ref{algo_switch}. For each recognized entity, we switch the recognized entity to a relevant but probably low-frequency entity with a probability of $\alpha$.
Such switching is divided into two types: with a probability of $\beta$, we switch the recognized entity to another one which the two entities have one of 13 relations of UMLS (other than \textbf{SR}); while  $1 -\beta$ is the probability of switching to an entity with only the \textbf{SR} relation, i.e. the two entities have the same concept ID in UMLS.


\begin{algorithm}[t]
\scriptsize
  \SetKwData{Left}{left}\SetKwData{This}{this}\SetKwData{Up}{up}
  \SetKwFunction{Union}{Union}\SetKwFunction{FindCompress}{FindCompress}
  \SetKwInOut{Input}{input}\SetKwInOut{Output}{output}
  \Input{An entity: $s$   \;
  Relation set in UMLS: $\mathcal{R}$ \;
  Switching probability distribution over relations $\mathcal{R}$: $p$\;
  Subject-object entity pairs associated to a specific relation:  $\Theta$ \;
  }
  \Output{A target switching entity $t$}
  \BlankLine
  \If{$\textrm{random}()<\alpha$}{
      \eIf{
            $\textrm{random}()<\beta$}
            {
                $r \leftarrow $ sample  a relation from $\mathcal{R}$ w.r.t.  $p$\;
                $t \leftarrow $ sample a object entity from $\Theta_r$ associated to the subject entity $s$ \;
            }
            {
                $t \leftarrow $ sample an entity that has the same concept ID with $s$\ (in \textbf{SR} relation);
            }
  }
\protect\caption{Entity Switching}
\label{algo_switch}
\end{algorithm}

We continue training a biomedical pre-trained Bert with more steps. Given a biomedical text sample, we first detect knowledge entities and follow the instructions in \S \ref{sec:entity switching} to generate a switched text.
We might switch multiple entities in a single text since there might be more than one recognized entity in it. 
Despite entity tokens might be masked, the predicted tokens are the replaced tokens after substitution instead of the original ones.


\subsection{Benefits of Entity Switching}

The benefits of entity switching are twofold. First,  \textbf{It augments training corpus with more low-frequency entities}. In general, one might get used to expressing a concept with his own preferences, even the concept (especially in biomedicine) could have different synonyms or have some low-level subclasses that also share its most features. These synonyms and homonyms might be lower frequency than the commonly-used concept and therefore under-represented in training corpus. By using entity switching, it could augment these under-represented concepts in the data side, while it does not change the model architecture. 

Secondly, \textbf{it aligns entities in relations}.  
Suppose we switch an entity on the context and it does not change the target predict words; this will lead to a consequence that the predictions of PLMs are invariant to entity switching, especially under polymorphism and synonymous substitution. A natural solution to such a consequence might be that the new entity will be converged to that of the switched entity during training, resulting in an alignment between them in the semantic space. 




\section{Experiments}

\subsection{Experimental Setup}
For continue pre-training, we use the PubMedDS dataset generated from \cite{medtype2020}, which is similar to the corpus used in PubMedBert \cite{gu2021domain} and all entities in the dataset are extracted and matched with the corresponding UMLS concept IDs. The dataset contains 13M documents and 44K concepts. To incorporate implicit knowledge into a PLM more efficiently, we randomly sampled 5 documents for each concept and finally retrieved 184,874 documents in total. 
We use BioBert, PubMedBert and Bio-LinkBert as our competitive baselines. See details in App.~\ref{app:baselines}.

\paragraph{Training Details} We continue training PubMedBert\cite{gu2021domain} with our method. Specifically, models are trained for 50 epochs on two tesla A100 for approximately 8 hours. The batch size is set to 64, with the AdamW\cite{loshchilov2017decoupled} as optimizer and a linear learning rate scheduler with a warm-up in 10\% of steps. In fine-tuning and knowledge probing, we follow the same method used in \cite{gu2021domain,sung2021can}.

\paragraph{Evaluation}
\textbf{BLURB} \cite{gu2021domain} evaluates language understanding and reasoning ability of models through 13 biomedical downstream tasks.
\textbf{BioLAMA}, a probing benchmark for probing language models in biomedical domain. To further probe hyponymous and synonymous knowledge in PLMs, we build two additional datasets \textbf{UMLS-Syn} and \textbf{UMLS-Hyp}. See details in App. \ref{app: additional probing data}.

\begin{table*}[htbp]
\centering
\vspace{-15pt}
\resizebox{\linewidth}{!}{
\addtolength\tabcolsep{-3.5pt} 
\begin{tabular}{l|ccccc|c|ccc|c|c|cc|l}
\toprule
 \textbf{Task} & \textbf{ \footnotesize BC5-chem} & \textbf{ \footnotesize BC5-disease}  & \textbf{ \footnotesize NCBI-disease} & \textbf{ \footnotesize BC2GM} & \textbf{ \footnotesize JNLPBA} & \textbf{\footnotesize EBM PICO} & \textbf{\footnotesize ChemProt} & \textbf{\footnotesize DDI}  & \textbf{\footnotesize GAD} & \textbf{\footnotesize BIOSSES} & \textbf{\footnotesize HoC} & \textbf{\footnotesize PubMedQA} & \textbf{\footnotesize BioASQ} & \textbf{\footnotesize BLURB score} \\
\hline
\textbf{PubMedBert} & 93.33 & \textbf{85.62} & 87.82 & \textbf{84.52} & 79.10 & 73.38 & 77.24 & 81.46 & 83.96 & 89.80 & 82.32 & 55.84 & 87.56 & 80.69 \quad\quad\quad \\
\hline
\textbf{Ours} & 93.11 & 85.19 & \textbf{88.65} & \textbf{84.52} & \textbf{79.47} & \textbf{74.55} & \textbf{77.58} & 82.18 & 83.93 & \textbf{92.86} & 84.68 & 67.4 & 92.14 & \textbf{83.21}  {\color{blue}\scriptsize $\uparrow _{3.12\%}$} \\
\textbf{Ours-w/o rel} & 92.93 & 85.21 & 88.08 & 83.81 & 79.29 & 73.54 & 77.43 & 80.73 & \textbf{84.07} & 92.04 & 84.86 & 63.4 & 92.14 & 82.47  {\color{blue}\scriptsize $\uparrow _{2.2\%}$} \\
\textbf{Ours-w/o syn} & 92.95 & 84.14 & 88.48 & 83.89 & 78.47 & 74.22 & 76.68 & 81.22 & 82.2 & 90.27 & \textbf{84.96} & \textbf{67.8} & 90.71 & 82.39  {\color{blue}\scriptsize $\uparrow _{2.1\%}$} \\
\textbf{Ours-w/o weak} & 93.37 & 85.09 & 87.51 & 84.35 & 78.97 & 73.27 & 76.43 & \textbf{82.23} & 82.19 & 90.39 & 84.74 & 66.6 & \textbf{93.57} & 82.44  {\color{blue}\scriptsize $\uparrow _{2.2\%}$} \\
\textbf{Ours-w useless} & \textbf{93.5} & 85.56 & 87.04 & 84.09 & 78.94 & 73.11 & 75.17 & 80.35 & 83.77 & 91.54 & 84.55 & 66.8 & 92.86 & 82.38  {\color{blue}\scriptsize $\uparrow _{2.1\%}$} \\
\textbf{Ours-w/o switch} & 92.79 & 84.5 & 87.98 & 83.81 & 79.11 & 73.28 & 76.58 & 80.74 & 82.15 & 88.15 & 84.22 & 57.8 & 88.57 & 80.72 {\color{blue}\scriptsize $\uparrow _{0\%}$} \\
\hline
\textbf{BioBert} & 92.85 & 84.70 & 89.13 & 83.82 & 78.55 & 73.18 & 76.14 & 80.88 & 82.36 & 89.52 & 81.54 & 60.24 & 84.14 & 80.34\quad\quad\quad \\
\textbf{BioLinkBert} & 93.04 & 84.82 & 88.27 & 84.41 & 79.06 & 73.59 & 77.05 & 81.14 & 82.98 & 93.63 & 83.37 & 65.2 & 91.43 & 82.54\quad\quad\quad \\

\bottomrule
\end{tabular}
}
\caption{
Evaluation on BLURB. The better-performing result on test set between  PubMedBert with and without the substitutions strategy is in \textbf{bold}. 
}
\label{blurb evaluation}
\end{table*}

\begin{table*}[htbp]
\centering \scriptsize
\addtolength\tabcolsep{-3.5pt} 
\begin{tabular}{c|c|ccccccc}
\toprule
Data & \textbf{Prompt} & \textbf{PubMedBert}   & \textbf{Ours} & \textbf{Ours-w/o rel} & \textbf{Ours-w/o syn} & \textbf{Ours-w/o weak} &  \textbf{Ours-w useless} &  \textbf{Ours-w/o switch} \\
\midrule
\multirow{2}{*}{UMLS} & Manual & 5.15/11.91 & \textbf{6.06}/\textbf{13.41} & 5.52/12.29 & 5.03/12.64 & 5.42/\underline{13.2} & \underline{6.05}/12.46 & 5.33/12.07\\

& Opti. & 12.33/27.4 & \underline{12.51}/\textbf{31.8} & 11.26/\underline{29.49} & 11.17/27.35 & 12.21/27.67 & \textbf{12.9}/28.32 & 12.37/27.2 \\

\midrule

\multirow{2}{*}{UMLS-Syn} & Manual & 4.61/10.88 & \textbf{5.59}/\textbf{12.3} & 5.18/11.8 & 4.69/11.71 & 5.05/\textbf{12.3} & \underline{5.48}/\underline{11.91} & 4.81/11.12\\

& Opti. & 10.69/23.57 & \underline{13.26}/\textbf{30.82} & 11.38/27.79 & 10.42/23.76 & 11.15/27.18 & \textbf{13.71}/\underline{30.19}  & 10.4/24.06 \\

\midrule

\multirow{2}{*}{UMLS-Hyp} & Manual & 4.3/11 & \textbf{5.01}/\textbf{11.68} & 4.85/11.47 & 4.42/11.25 & 4.61/11.56 & \underline{4.92}/\underline{11.7} & 4.3/11.0 \\

& Opti. & 10.4/23.75 & \underline{12.21}/\textbf{29.62} & 11.71/25.52 & 10.74/23.47 & 10.9/24.78 & \textbf{12.52}/\underline{26.7} & 10.96/24.78\\

\bottomrule
\end{tabular}
\caption{\label{BioLAMA}
Knowledge probing on BioLAMA with manual prompt and OptiPromp\cite{zhong2021factual} method. Acc@1/Acc@5 of each model are reported. The highest and the second highest accuracy is in \textbf{bold} and \underline{underlined}.
}
\end{table*}

\subsection{Experimental Setting}



Based on the similarity of the corresponding relations, relations in UMLS were classified into three classes: 1) \textit{strong} similarity relations including \textbf{CHD}, \textbf{RN}, and \textbf{RL}, denoted as $\mathcal{R}_1$; 2) \textit{weak} similarity relations \textbf{PAR}, \textbf{RB}, and \textbf{RQ}, denoted as $\mathcal{R}_2$; and 3) other relations, denoted as $\mathcal{R}_3$. For ablation study of polymorphism, synonyms and switching method, we constructed multiple configurations shown in Tab.~\ref{tab:configuration setting}.

\begin{table}[t]
\tiny
\vspace{-5pt}
    \centering
    \begin{tabular}{lllll}
        \toprule
        Name & $\alpha$ & $\beta$ & $p$ & $\mathcal{R}$ \\
         \midrule
         Ours           &    0.2    &   0.8    &   5:1:0&      $\mathcal{R}_{1} \cup \mathcal{R}_{2}$ \\
         Ours-w/o rel       &    0.2    &   0      &   None &      $\emptyset$       \\
         Ours-w/o syn      &    0.2    &   1      &   5:1:0&     
         $ \mathcal{R}_{1} \cup \mathcal{R}_{2}$         \\
         Ours-w/o weak &    0.2    &   0.8    &   1:0:0&
         $  \mathcal{R}_{1}$\\
         Ours-w useless       &    0.2    &   0.8    &   1:1:1&     $ \mathcal{R}_1 \cup \mathcal{R}_2  \cup \mathcal{R}_3 $         \\ 
         Ours-w/o switch       &    0    &   0    &   None &     $\emptyset$         \\ 
         \bottomrule
    \end{tabular}
    \caption{Settings of different configuration.
    }
    \label{tab:configuration setting}
\end{table}

\subsection{Experiment Results}
\label{sec:blurb}


Evaluation on BLURB is shown in Tab.~\ref{blurb evaluation}. Compared to PubMedBert, our model significantly outperformed the baseline PubMedBert  with 3.12\% improvement.
This is due to that our models could better capture knowledge that is encapsulated in low-frequency entities; which could be augmented by entity switching.
Specifically, our standard model outperformed BioLinkBERT which was the state-of-the-art model in BLURB benchmark,  demonstrating the effectiveness of our approach.

As an ablation study, both polymorphism and synonymous substitution are beneficial to our approach, see \textbf{Ours-w/o syn} and  \textbf{Ours-w/o rel} respectively. Interestingly, using useless relations (i.e., $\mathcal{R}$) seems harmful (see the comparison between \textbf{Ours-w useless} and \textbf{Ours}). The well-designed configuration (\textbf{ours}) that leverages more strong relations and a few weak relations achieves the best performance.

\subsection{Study on Knowledge Probing}


Evaluation on BioLAMA is shown in Tab.~\ref{BioLAMA}. Our model achieved the best results in knowledge probing in UMLS, demonstrates effectiveness of entity switching which  successfully injects some UMLS knowledge into the model. When replacing subjects with synonyms and hyponyms (UMLS-Syn and UMLS-Hyp benchmarks), baseline models showed a significant performance drop while the drop our models is relatively negligible. This demonstrates that our models could better capture hyponyms and synonyms.

We found that switching with either synonymous substitution  or polymorphism achieved better performance than that without switching, suggesting that switching entities to polymorphic and synonymous entities enhances the knowledge ability of  models. 
Interestingly, both too big or too small switching probabilities for switching weak relations will lead to worse performance. A moderate probability for switching weak  relations performs the best, since we have to trade off between the switching scale and noises; switching in weak relations could inject more knowledge in PLMs but these knowledge might be noisy. The findings are generally consistent to \S\ref{sec:blurb}.


\section{Conclusion}
Through our observations, we found that the concepts of UMLS are polymorphic and that the distribution of entities and knowledge in the training corpus is usually long-tailed. We therefore propose a new knowledge injection method that increases the probability of occurrence of low-frequency entities and implicitly injects UMLS knowledge by replacing entities with different probabilities in the corpus. Our experimental results demonstrate that we successfully inject more knowledge into the model and exceed the performance of baselines on various biomedical downstream tasks.

\section*{Limitations}
While it is an effective way to incorporate knowledge into PLMs, it needs multiple epochs to align entities which makes it difficult to train a PLM from scratch with our method.

\section*{Ethics Statement}
There are no ethics-related issues in this paper. The data and other related resources in this work are  open-source and commonly-used by many existing work. 

\bibliography{anthology,custom}
\bibliographystyle{acl_natbib}

\appendix


\section{ Reasons to  select biomedical scenario}
The reasons to select the biomedical scenario are manyfold:
\begin{itemize}
    \item biomedical knowledge bases are typically more knowledge-intensive than general domain; 
    \item it contains more low-frequency knowledge  where PLMs usually fail to capture;
    \item there are some well-designed relational knowledge bases, e.g.,  Unified Medical Language System (UMLS)\cite{bodenreider2004unified} which consists of more than 4M concepts and 900 relations.
\end{itemize}


\section{An example of report bias}
\label{sec:example_complementary}
\begin{figure*}[ht]
\centering \small
\includegraphics[width=\textwidth]{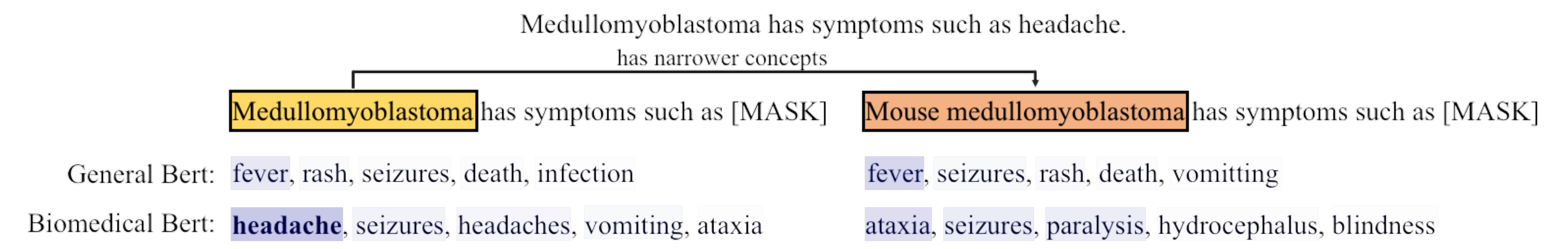}
\caption{A case of pilot study. Predicted tokens of masked location are shown. The gradient of blue indicates the prediction probability of the token in language models.}
\label{pilot_study_fig}
\end{figure*}

 As shown in Fig.~\ref{pilot_study_fig}, biomedical knowledge cannot be well-captured by a general Bert (see the left) since biomedical entities are relatively low-frequency in general corpora on which Bert is trained. 
When we replace a concept with a narrower ({\tt Mouse medulloblastoma } is a special case of  {\tt medulloblastoma}, see the right part), both Bert and its biomedically-adapted one fail to predict the masked word since the narrower concepts are usually more low-frequency.

\section{Baseline Models}
\label{app:baselines}

\begin{itemize}
[noitemsep,nolistsep,leftmargin=*]
    \item \textbf{BioBert} \cite{lee2020biobert}: A widely used biomedical Bert continue trained on PubMed data.
    \item \textbf{PubMedBert} \cite{gu2021domain}: Biomedical Bert pre-trained on PubMed data from scratch.
    \item \textbf{Bio-LinkBert} \cite{yasunaga2022linkbert}: A state-of-the-art pre-train model in BLURB benchmark, it is continue trained on PubMedBert with a sentence level auxiliary task.
\end{itemize}

\section{Additional Probing Data}
\label{app: additional probing data}
To probe synonymous and hyponymous knowledge in PLMs, we further constructed two new datasets. Based on the original BioLAMA dataset \textbf{UMLS-Syn} and \textbf{UMLS-Hyp}. For each piece of original data, we constructed two pieces of data of which the subjects are replaced with corresponding hyponyms or synonyms. Finally, we retrieved 18447 pieces of data in \textbf{UMLS-Syn} and 10978 pieces in \textbf{UMLS-Hyp}. 

\section{Relation Details}
\label{app:Relation Details}
Show in Tab.~\ref{Relation description}.
\begin{table}[htb]
\centering \small
\resizebox{\linewidth}{!}{
\begin{tabular}{l|l}
\toprule[2pt]
\textbf{Relation} & \makecell[l]{Description}\\
\midrule[1pt]
AQ & \makecell[l]{Allowed qualifier.} \\
\midrule
CHD & \makecell[l]{has child relationship in a Metathesaurus source vocabulary.}\\
\midrule
DEL & \makecell[l]{Deleted concept.} \\ 
\midrule
PAR & \makecell[l]{Has parent relationship in a Metathesaurus source vocabulary.} \\ 
\midrule[1pt]
QB & \makecell[l]{Can be qualified by.} \\
\midrule
RB & \makecell[l]{Has a broader relationship.}  \\ 
\midrule
RL & \makecell[l]{The relationship is similar or "alike". The two concepts are \\ similar or "alike". In the current edition of the Metathesaurus,\\ most relationships with this attribute are mappings provided\\ by a source, named in SAB and SL; hence concepts linked by\\ this relationship may be synonymous, i.e. self-referential:\\ CUI1 = CUI2. In previous releases, some MeSH Supplementary\\ Concept relationships were represented in this way.}  \\
\midrule
RN & \makecell[l]{Has a narrower relationship.} \\ 
\midrule
RO & \makecell[l]{Has relationship other than synonymous, narrower, or broader.} \\
\midrule
RQ & \makecell[l]{Related and possibly synonymous.}  \\ 
\midrule
RU & \makecell[l]{Related, unspecified.}  \\
\midrule
SY & \makecell[l]{source asserted synonymy.} \\
\midrule
XR & \makecell[l]{Not related, no mapping.}\\
\bottomrule[2pt]
\end{tabular}
}
\caption{
Description of relations in UMLS.
}
\label{Relation description}
\end{table}

\section{Related Work}

\textbf{Biomedical domain adaption of PLM}
In order to build a PLM that better understands biomedical texts and performs better in biomedical downstream tasks, \cite{lee2020biobert,peng2019transfer} continue pre-trained a general-domain Bert with additional corpus from Pubmed and MIMIC\cite{johnson2016mimic}. The additional training steps enable the model to store more biomedical-related knowledge and to improve significantly on NLP tasks in this area. This also shows that the performance of the pre-trained model depends heavily on the distribution of implicit knowledge in the corpus. Furthermore, \cite{gu2021domain,beltagy2019scibert} pre-trained domain-specific models from scratch with abundant unlabelled corpus, their results show that pre-training language models from scratch results in substantial gains over continual pre-training of general-domain language models. This reinforces that the knowledge stored in the model parameter depends on the distribution of implicit knowledge in the training corpus.

\textbf{Incorporate structural knowledge in PLMs}
To incorporate structural knowledge, \cite{yuan2021improving, peters2019knowledge} first encode the structured knowledge graph and then fuse the encoding of the entities with the text encoding in Transformer. However, the graph representation space and the text space are difficult to integrate. \cite{liu2020self} incorporates synonym knowledge into the model by bringing synonyms in the semantic space closer together and distancing non-synonyms based on UMLS synonyms. But we believe that the representation of entities should not only be related to their own representations but should also depend on their corresponding contexts.

\end{document}